\documentclass[final,conference]{IEEEtran}
\usepackage{soul}
\usepackage{graphicx,array}
\usepackage{cite}
\usepackage{amsmath,amssymb,amsfonts}
\usepackage{graphicx}
\usepackage{textcomp}
\usepackage{algorithm}
\usepackage{algpseudocode}
\usepackage{color}

\usepackage{verbatim}
\usepackage[breaklinks=true]{hyperref}
\usepackage{cite}
\usepackage{textcomp}
\usepackage[table]{xcolor}
\usepackage{enumitem}
\usepackage{multicol}
\usepackage{multirow}
\usepackage{gensymb}
\usepackage{todonotes}

\usepackage{algorithm}
\usepackage{algpseudocode}

\usepackage{bm}

\usepackage{booktabs,tabularx,enumitem,ragged2e}
\usepackage{colortbl}
\usepackage[utf8]{inputenc}

\usepackage{xcolor}

\usepackage{array}
\usepackage{multirow}
\usepackage{graphicx,array}
\usepackage[table]{xcolor}

\usepackage[singlelinecheck=false]{caption}

\usepackage{romannum}

\usepackage{caption}
\usepackage{subcaption}

\begin{document}

\title{TwinExplainer: Explaining Predictions of an Automotive Digital Twin}

\author{\IEEEauthorblockN{Subash Neupane}
\IEEEauthorblockA{\textit{Computer Science \& Engineering} \\
\textit{Mississippi State University}\\
Starkville, MS, USA \\
sn922@msstate.edu}
\and
\IEEEauthorblockN{Ivan A. Fernandez}
\IEEEauthorblockA{\textit{Computer Science \& Engineering} \\
\textit{Mississippi State University}\\
Starkville, MS, USA \\
iaf28@msstate.edu}
\and
\IEEEauthorblockN{Wilson Patterson}
\IEEEauthorblockA{\textit{Computer Science \& Engineering} \\
\textit{Mississippi State University}\\
Starkville, MS, USA \\
wep104@msstate.edu}
\and
\IEEEauthorblockN{Sudip Mittal}
\IEEEauthorblockA{\textit{Computer Science \& Engineering} \\
\textit{Mississippi State University}\\
Starkville, MS, USA \\
mittal@cse.msstate.edu}
\and
\IEEEauthorblockN{Milan Parmar}
\IEEEauthorblockA{\textit{Computer Science \& Engineering} \\
\textit{Mississippi State University}\\
Starkville, MS, USA \\
parmar@cse.msstate.edu}
\and
\IEEEauthorblockN{Shahram Rahimi}
\IEEEauthorblockA{\textit{Computer Science \& Engineering} \\
\textit{Mississippi State University}\\
Starkville, MS, USA \\
rahimi@cse.msstate.edu}
}


\maketitle
\begin{abstract} Vehicles are complex Cyber Physical Systems (CPS) that operate in a variety of environments, and the likelihood of failure of one or more subsystems, such as the engine, transmission, brakes, and fuel, 
can result in unscheduled downtime and incur high maintenance or repair costs. 
In order to prevent these issues, it is crucial to continuously monitor the health of various subsystems and identify abnormal sensor channel behavior.
Data-driven Digital Twin (DT) systems are capable of such a task. 
Current DT technologies utilize various Deep Learning (DL) techniques 
that are constrained by the lack of justification or explanation for their predictions. 
This inability of these opaque systems can influence decision-making and raises user trust concerns. 
This paper presents a solution to this issue, where the \emph{TwinExplainer} system, with its three-layered  architectural pipeline, explains the predictions of an automotive DT. Such a system can assist automotive stakeholders in understanding the global scale of the sensor channels and how they contribute towards generic DT predictions. TwinExplainer can also visualize explanations for both normal and abnormal local predictions computed by the DT.


\end{abstract}


%
\IEEEpeerreviewmaketitle

\section{Introduction}
Modern vehicles, with their array of sensors, can be thought of as a subset of the broader category of Cyber-Physical Systems (CPS), which as a whole generates a volume of operational data. 
Through the usage of current CPUs and GPUs, we now have access to unprecedented computational power. 
These big vehicular datasets can be utilized in tandem with Artificial Intelligence (AI) and Machine Learning (ML) technologies operating on this hardware to construct data-driven intelligent systems \cite{neupane2022temporal}. 
One such intelligent system is a Digital Twin (DT). 
As can be deduced from its etymology, a Digital Twin (DT) is defined as ‘an integrated multi-physics, multiscale, probabilistic simulation of a complex product, which functions to mirror the life of its corresponding twin’ \cite{glaessgen2012digital}.
The primary goal of this technology is to function as a panoptic mirror of a physical body in the digital world.
This has many intrinsic advantages since the physical object may adapt to change its real-time behavior in response to the input given by the digital twin.
Conversely, the bridging enables the simulation to be able to precisely mimic the real-world condition of the physical body. 

To realize a full cyber-physical system, the digital twin must be dynamically coupled with the physical model, which may be accomplished via real-time sensory data.
In simple terms, DT is a live (virtual) model of a physical object that utilizes real-time sensor data to simulate behavior, monitor activities, and predict the future of physical assets.
The DT technology employs machine learning algorithms to analyze and identify data patterns and sensor channel activity from enormous volumes of sensor data. 
AI and ML yield useful insights into data for optimizing performance, maintaining components, reducing emissions, and increasing efficiency \cite{awsdt}.


One of the major drawbacks of employing deep learning models is that they are regarded as black boxes with opaque decision algorithms. 
Because of their nested and non-linear structure, these systems, which constitute the vast majority of the state-of-the-art, may attain outstanding prediction accuracy. However, they are constrained by a lack of justification for their predictions \cite{samek2017explainable}. 
In contrast, the inability of an opaque model to identify and explain the precise information in data that drives their prediction, causes user trust concerns \cite{marshan2021artificial}. 
Therefore, if these systems are to be utilized in automotive domains, some degree of explanation of their decision process must be possible.

To address these issues, and in the spirit of developing DTs that are interpretable to automotive stakeholders, we present \textit{TwinExplainer}, which approximates and explains DT predictions by utilizing SHapley Additive exPlanations (SHAP) \cite{lundberg2017unified} as an explainable algorithm. 
There is no one answer to the back-box interpretability issue at the moment. Nonetheless, a slew of possible explanations have emerged, exploring, and exploiting various aspects of the machine learning process in order to provide explanations for opaque models \cite{9927396}. A feature-based method (for example, SHAP) is one of these possible explanations that employ features as its explanation medium. The goal of this method is to determine how much each feature is responsible for the output predictions. 

In this paper, we expand our previous work \cite{neupane2022temporal} that predicts the anomalous sensor channel behaviors in various Functional Working Groups (FWG) also known as subsystems, of automotive systems, where we treat the combination of four major FWGs such as engine, transmission, fuel, and braking subsystems as a DT system. 
{In this particular DT system, we make use of the Temporal Convolution Network (TCN) as our deep learning classifier in order to monitor and forecast the behaviors of the sensor channels.} 
The dataset used in this experiment is the Vehicle Performance, Reliability, and Operations (VePro) dataset that was created by the United States (U.S.) Army Corps of Engineers and Mississippi State University (MSU), which contains operational data from a fleet of military vehicles, such as time series sensor measurements, detected faults, and maintenance reports. We utilize a feature-based approach to explain the prediction made by the TCN model for this paper.

The major contributions of this paper are:

\begin{itemize}
    \item We create a data-driven framework called \emph{TwinExplainer} that generates explanations from an automotive Digital Twin (DT).
    \item Typically, mechanics, operators, etc. cannot comprehend the predictions made by predictive DT models that utilize deep learning architecture. We demonstrate that it is possible to generate explanations for such models, which in turn augments user trust and decision-making.
    \item TwinExplainer creates visualizations that explain the predictions made by a DT. Focusing on features, we explain the prediction by demonstrating feature attribution and importance in relation to a particular prediction (local explanation) and the system as a whole (global explanation).
\end{itemize}

The rest of the paper is organized as follows: Section \ref{background} discusses the background and related work. 
Section \ref{architecture} discusses the \emph{TwinExplainer} architecture. 
In Section \ref{experiment} we discuss our experiment and methodology. Section \ref{explanations}  talks about generating explanations. Finally, Section \ref{conclusion}  concludes the paper.

\section{Background and Related Works}
\label{background}
In this section, we present some related work on Digital Twins (DT), Predictive Maintenance (PdM), Explainable Artificial Intelligence (XAI), and Explainable Digital Twins (X-DT).

\subsection{Digital Twin (DT)}

The concept of using ``physical twins”, a precursor to Digital Twins (DTs), is not new. This concept first appeared in the 1970s during NASA's Apollo program. 
Grieves et al. introduced DT informally in 2002 and subsequently formalized the concept in their white paper \cite{grieves2014digital}.
{In essence, DT is a virtual prototype of physical assets that simulates real-time operational conditions in order to behave like a real physical asset.}


DT technology has benefited a wide array of industries such as healthcare \cite{liu2019novel, elayan2021digital}, aviation \cite{shafto2012modeling, tuegel2011reengineering, liu2021digital}, manufacturing \cite{banerjee2017generating, tao2018digital, neupane2022temporal}, cyber-physical systems \cite{alam2017c2ps, tao2019digital}, security \cite{ables2022creating}, by providing a diverse range of applications. In the automotive industry, many functions of DT are being utilized, particularly in vehicle development, operation, and maintenance. Manufacturers have successfully integrated DT in vehicle development process in order to lower development costs. Developers can now create a high number of digital prototypes, allowing them to simulate countless scenarios for vehicles before they are ever physically created \cite{biesinger2019facets}. Other vehicular applications for DT provide numerous opportunities during a vehicle’s operating life. Sensors installed on vehicles are capable of capturing operating data and updating a vehicle’s digital twin throughout its lifetime. The captured data provides feedback to vehicles that are on the road, allowing for real-time diagnostics and assessments. 

\subsection{Predictive Maintenance (PdM)}
Maintenance is a crucial process for maintaining a system operating in its intended mode, whether through rectifying problems or adopting preventive actions \cite{theissler2021predictive}. Maintenance strategies are divided into three broad categories in the body of literature, such as Reactive Maintenance (RM) \cite{mobley2002introduction}, Preventive Maintenance (PM) \cite{mobley2002introduction, wan2017manufacturing}, and Predictive Maintenance (PdM). The first strategy, also known as corrective maintenance, is a ``run-to-failure" method that aims to repair systems after a failure has occurred. Conversely, PM usually referred to as scheduled maintenance, plans maintenance operations using fixed-time intervals to reduce the likelihood of failures. While the third category of PdM, also known as Condition-Based Monitoring (CBM) \cite{williams1994condition}, aims to predict when the equipment is likely to fail based on the system’s health, operation, and maintenance data.

PdM leverages the data often collected from embedded sensors in systems and components for predictive analysis using predictive algorithms. These models can effectively discover behavioral trends and patterns in operational data, a task closely related to the modeling system's normal behavior, and detect deviations, referred to as anomalies, which may indicate a current or impending failure \cite{theissler2021predictive, ran2019survey}. An anomaly is a deviation or potential error within the taxonomy of error, fault, and failure, where an error is caused by a fault and may result in a failure. The process of identifying these errors is known as ``anomaly detection" and can consequently be used for PdM tasks \cite{theissler2021predictive}. 
In the following subsections, we discuss explainable methods.

\subsection{Explainable Artificial Intelligence (XAI)}
In recent years, interest in the explainability and interoperability of AI, also known as ``eXplainable Artificial Intelligence" (XAI), has significantly increased. The empirical success of deep learning models has resulted in the emergence of opaque decision systems, 
which are {often} complex black-box models {to the end users.} 
Explanations supporting the output of a model are crucial for providing AI stakeholders with transparency regarding predictions \cite{zhang2022explainable, 9927396}. Explanation pertains to the manner in which intelligent systems justify their decisions. 
\emph{Scope of explainability} and \emph{model dependence} is a common category found in the literature concerning the taxonomy of explainability \cite{9927396}. These categories are elaborated upon in the subsections that follow. 

\subsubsection{Local Explainability}
Explainability at the local level refers to a system's ability to communicate to a user the reasoning behind a specific decision. 
Local Interpretable Model Agnostic Explanation (LIME) \cite{ribeiro2016should}, the Anchors \cite{ribeiro2018anchors} and the Leave One Covariate Out (LOCO) \cite{lei2018distribution} are examples of local explanation techniques. Individual predictions are interpreted by LIME using a local surrogate model. Another important explainer module is Shapely Additive Explanation (SHAP), proposed by Lundberg et al. \cite{lundberg2017unified}. It determines the significance of each feature for each prediction.

\subsubsection{Global Explainability}
The global explainability of a model makes it simpler to comprehend the logic underlying all potential outcomes. These models enable an understanding of the attributions for various input data and shed light on the general decision-making process of the model. Examples of global explainer modules include Submodular Pick-LIME (SP-LIME), an extension of the LIME module, Concept Activation Vectors (CAVs) \cite{kim2018interpretability}, Global Interpretation via Recursive Partitioning (GIRP) \cite{yang2018global}, etc. 

\subsubsection{Model-agnostic}
Model-agnostic methodologies are not bound to any particular class of ML model. They are therefore modular in the sense that the explanatory module is independent of the model for which it generates explanations. Visualization, knowledge extraction, influence methods, and example-based explanations are some examples of model-agnostic interpretability techniques \cite{adadi2018peeking}.
\subsubsection{Model-Specific}
Model-specific explainer approaches are designed exclusively for a particular model type. Model-specific interpretations include, for instance, the interpretation of regression weights in a linear model. In this sense, the interpretation of intrinsically interpretable models, such as regression, rule-based models, and decision trees, is always model-specific. 

\begin{figure}[h]
    \centering
    \includegraphics[scale=.77]{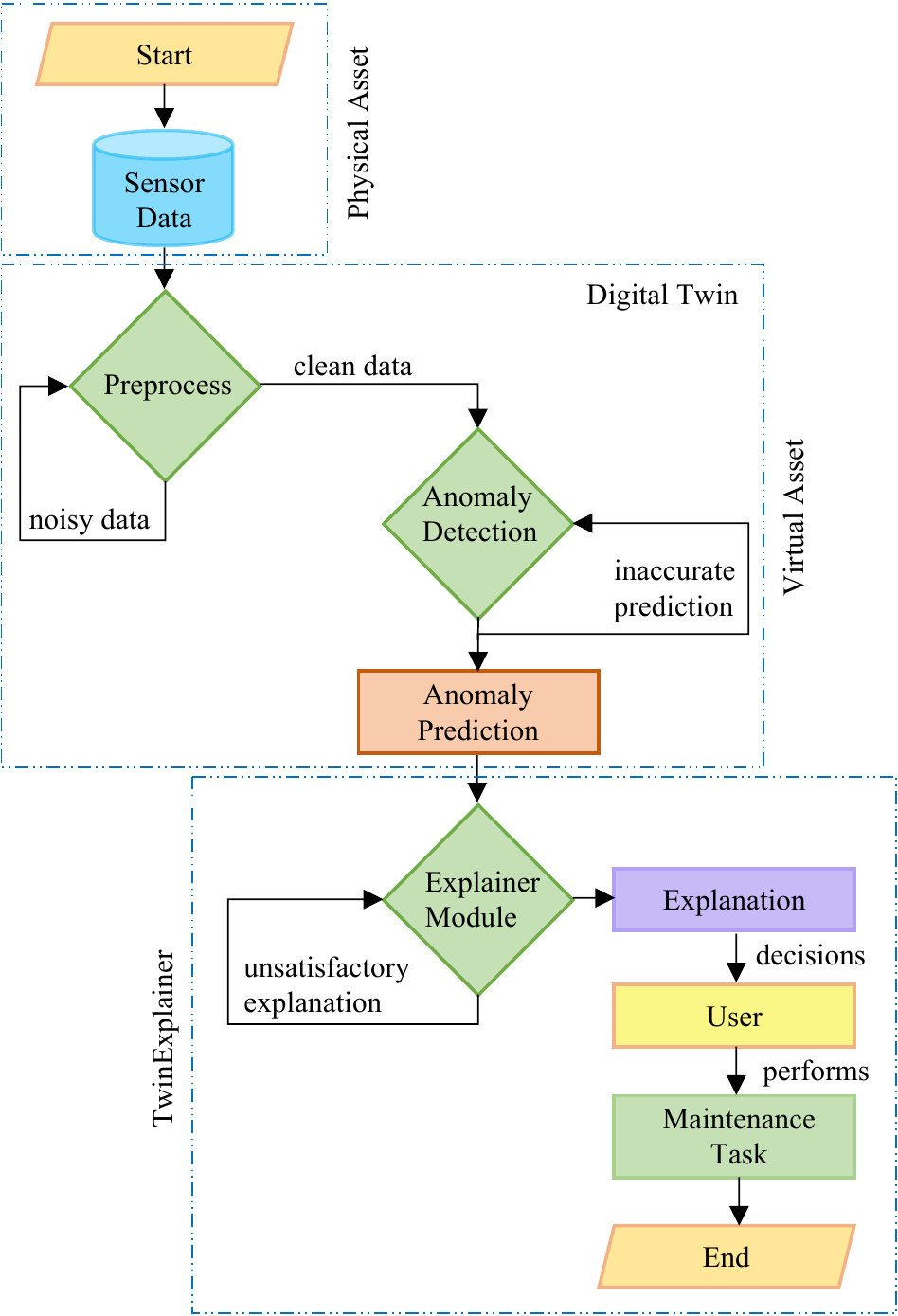}
    \caption{{Process flow diagram for the \emph{TwinExplainer} model. Sensor data collected from physical assets, such as vehicles, serves as the model's initial input. After preprocessing the raw sensor data, it is fed into an anomaly detection DT model, which forecasts future time-series predictions. The explainer module receives a prediction and generates an explanation. Based on the explanation of the prediction, a user can decide to schedule a maintenance event.} 
    }
        \label{fig: Flow_chart}
\end{figure} 

\subsection{Explainable Digital Twins (X-DT)}
In the context of DT enabled by AI/ML/DL classifiers, decisions can be a product of autonomous reasoning processes and may require explanations. {Explaining classifier predictions will assist different stakeholders (such as developers, operators, mechanics, etc.) in making decisions, as it enables them to comprehend the context-based justification for various computed decisions} \cite{zhang2022explainable}.


Numerous studies have been presented in which DTs are discussed in terms of their interpretable decision-making. In \cite{rao2019digital} a risk diagnosis DT system is proposed to enhance the decision-making for liver diseases with XAI. In doing so, the authors used the state-of-the-art Local Interpretable Model-Agnostic Explanations (LIME) methods to approximate the decision made by the Random Forest classifier. Explanations for dynamically selecting a DT model from a library of components-based reduced order models based on sensor measurements of {an} {Unmanned Aerial Vehicles (UAV)} is provided in \cite{kapteyn2020toward}. They investigate several regions of damage of varying degrees of severity across the aircraft wings. 
{In another work, Zhang et al.} \cite{zhang2022explainable} {investigate the problem of explainability with a human-in-the-loop approach for dynamic data-driven DTs where the authors propose} an architecture to determine when and where explanations can support data-driven or system-calculated decisions. Similarly, Wang et al. \cite{wang2021explainable} in their paper, proposed a framework of explainable modeling to enable the collaboration and interaction between developers and other stakeholders of the DT ecosystem. The authors use a DT factory use case to show how three explainability scores (model-based, scenario-based, and goal-oriented) can be used to measure the value of modeling that is easy to understand.

{Next, we describe data flow and the architecture of the \emph{TwinExplainer} system.}

\section{TwinExplainer Architecture}
\label{architecture}

\begin{figure*}[h]
    \centering
    \includegraphics[scale=.85]{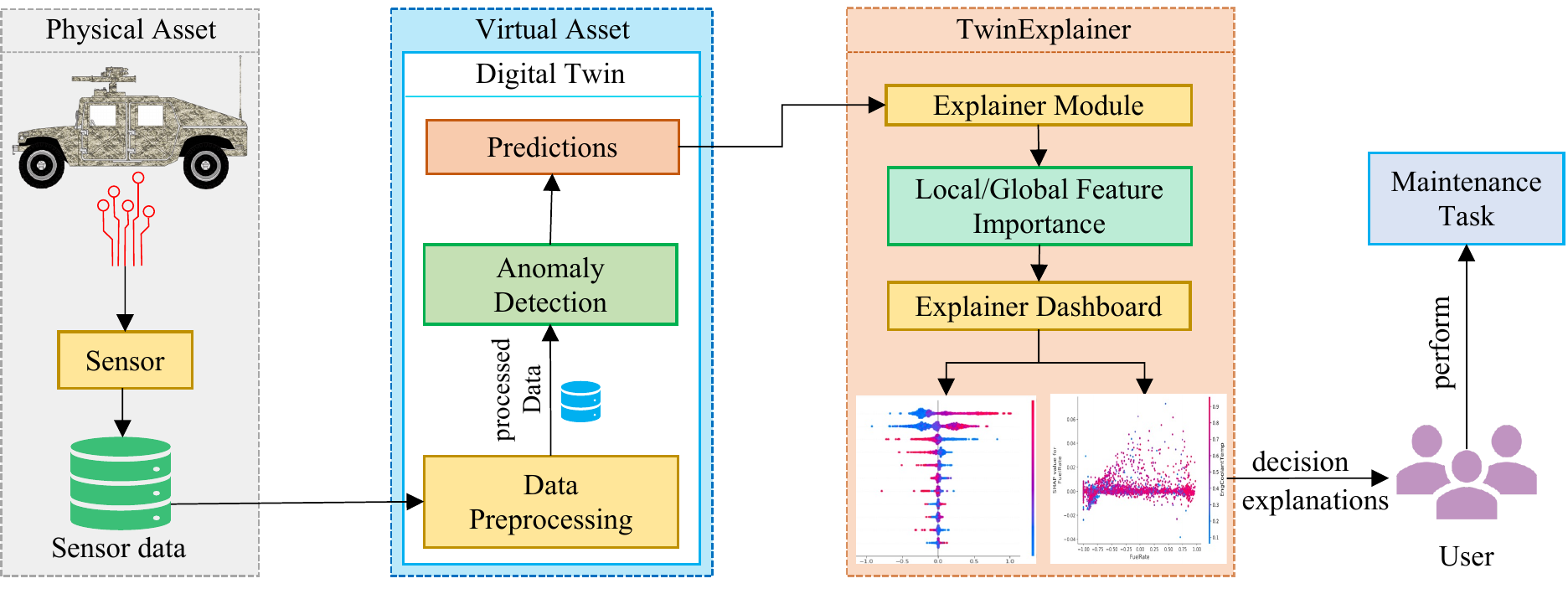}
    \caption{A graphical representation of the TwinExpainer's three-phase architecture. Raw sensor data is collected and stored during the physical asset phase. The data is then cleansed, transformed, and scaled in the digital twin stage. This phase generates high-quality data, which is then passed to a digital twin module, which predicts anomalous data patterns. The explainer module takes the prediction as input in the third phase and generates an explanation using visualization techniques to help the user understand why DT made particular prediction. 
    }
        \label{fig: system_architecture}
\end{figure*} 

This section describes the architecture of a deep learning-enabled digital twin system with XAI integration, referred to as \textit{TwinExplainer}. The majority of related work involving vehicular digital twins focuses on improving prediction accuracy, while ignoring the interpretability of such predictive outcomes. In this paper, we address the issue of the \textit{interpretability} of {predicted outcomes of} digital twin systems. {The goal of the \emph{TwinExplainer} system is to aid an automotive mechanic in comprehending the computed prediction decision by a DT via various visualization techniques and user trust augmentation in AI systems.}

In Fig \ref{fig: Flow_chart}, we summarize the {data flow} in {the} \emph{TwinExplainer} system. The sensor data is first preprocessed, to remove noise, incomplete data, and redundant observations. It is also normalized and transformed. This process is repeated until we have a clean data subset. 
Following that, high-quality, cleaned data is fed into a DT module, which forecasts sensor channel readings and flags anomalous data instances.
The process is repeated in cases of incorrect anomaly prediction by the DT. The predictions are then fed into an explainer module, which generates explanations. If an unsatisfactory explanation is provided, the process is repeated to meet various stakeholders' explainability needs. For example, a set of explanations might be too complicated for an automotive operator. They may request additional explanations that are easier to understand or resources that can teach them more about the topic. In such a case, developers can modify the explainer module to accommodate the user's requirements \cite{9927396}. 
{Finally, based on the explanation provided by \emph{TwinExplainer}, the user can make an informed decision on fault diagnosis, prognostics, and repairs, which mitigates the risk of abrupt component failure to some degree.}

Figure \ref{fig: system_architecture}, depicts the three phases of the \emph{TwinExplainer} architecture: physical assets, virtual assets, and the twin explainer. 
These various phases comprise our explanation framework. The rest of this section describes each phase in detail and elaborates on our techniques.

\subsection{Physical Asset}
Real-world attributes, such as objects, equipment, systems, and components, constitute the physical asset. In the context of vehicular systems, sensors and actuators are the two primary physical components that are coupled. The sensor detects a physical condition and transmits an electronic signal to the Electronic Control Unit (ECU) based on the detected and measured characteristics. Actuators, on the other hand, are physical components that are primarily responsible for moving and regulating a mechanism in order to complete a task. {Actuators can, for example, modify engine idle speed and adjust camshaft valve opening and closing to optimize engine performance and fuel economy based on the sensor signals.} 

Modern automobiles are equipped with several sensors that generate a large volume of operational data. This data is expressed as channels, such as \emph{vehicle speed, engine speed, engine percent torque, brake switch, accelerator pedal position, fuel rate, engine coolant temperature, transmission oil temperature, selected gear, injector control pressure}, etc. Typically, this data is collected in a time-series format.

During this phase, multichannel sensor data from multiple embedded sensors is collected and stored in a datastore. 
One can leverage these datasets to build intelligent systems, such as a Digital Twin (DT) with deep learning classifiers, that are capable of investigating, analyzing, and monitoring a subsystem or an overall automotive system. For instance, one can develop a DT for the fuel subsystem that can foresee and predict the behavior of fuel system-related channels. This would enable stakeholders in formulating a sensible maintenance plan, {repairs, fault diagnosis, and prognostics} to limit the possibility of unplanned operating interruptions in the physical world \cite{piromalis2022digital}.



\subsection{Virtual Asset} \label{virtual_asset_digital_twin}

In the automotive sector, DTs are often utilized for the complete or partial development of a virtual vehicular model. To model a DT system and evaluate the vehicle\textquotesingle s true performance, it is necessary to collect data on its operations, behaviors, and functions. \emph{TwinExplainer} employs the Vehicle Performance, Reliability, and Operations (VePRO) dataset, \cite{neupane2022temporal} which comprises operational sensor data from a fleet of military vehicles to construct a vehicular DT (See Section \ref{Modeling_dt} for more information in DT modeling). {More details about the dataset are available in Section} \ref{dataset} and Table \ref{table:features_description} showcases a summary of different channels present in the VePRO dataset.



The first stage in this phase is data processing, in which we remove inaccurate, incomplete, and duplicated observations. The data is then scaled and transformed to produce a high-quality data subset. This dataset is then fed into the deep learning-based anomaly detection model. The model performs time-series forecasting on connected sensor channels as part of its prediction. The predicted and actual observations are then compared; if the actual observations exceed the pre-defined threshold, the system marks that observation as an anomaly. In this paper, we employ the same anomaly detection system that we created in our previous work \cite{neupane2022temporal}.

\begin{table*}[ht]
{\renewcommand{\arraystretch}{1.20}%
\begin{tabular}{|p{1.5cm}|p{2cm}|p{2.3cm}|p{2.7cm}|p{7.3cm}|}
\hline
\rowcolor{lightgray!20!}
FWG         & Sensor Channel                    & Full name                                     & Unit                           & Description                                                                        \\ \hline
Time         & UTC\_1HZ                   & 1 Hz Data                                    & Seconds                            & Time series data from the 1 Hz channels.                                       \\ \hline

                                    & EngCoolantTemp             & Engine Coolant Temperature                   & Degree Fahrenheit (°F)            & Captures how heavy the engine is working and correlates well   with the fuel rate. \\\cline{2-5}  
                                    & PctEngLoad                 & Percentage Engine Load                       & Percent (\%)                             & Amount of load required for the engine to perform driving.                         \\\cline{2-5}  
\multirow{6}{*}{Engine}              & EngPctTorq                 & Engine Percentage Torque                     & Percent (\%)                            & The calculated output torque of the engine.                           \\\cline{2-5}  
                                    & BoostPres                  & Turbo Boost Pressure                         & Pound per square inch (psi)              & Indicates the air pressure information at any given moment.                        \\\cline{2-5}  
                                     & AccelPedalPos              & Acclerator Pedal Position                    & Percent (\%)                             & Captures acceleration, deceleration, and steady state   condition.                 \\\cline{2-5}  
                                    & IntManfTemp                & Intake Manifold Temperature                  & Degree Fahrenheit (°F)                  & Indicates the temperature of the air inside the intake   manifold.                 \\\cline{2-5}  \hline
                                    

& VehSpeedEng                & Vehicle Speed                                & Miles Per Hour (m/s)                   & Speed of vehicle -  major contributor to the fuel consumption.                      \\\cline{2-5} 
             & TransOilTemp               & Transmission Oil Temperature                 & Degree Fahrenheit (°F)                   & Captures temperature in all operating condition.                                   \\\cline{2-5} 
\multirow{6}{*} {Transmission} & TrSelGr                    & Transmission Selected Gear                   & Unitless                                 & Represents which transmission gear the vehicle is currently in.                  \\\cline{2-5}  
             & TransTorqConv
             LockupEngaged & Transmission Torque Converter Lockup Engaged & Base 10 Integer Number                  & Captures whether or not the torque converter lockup is   engaged.                 \\\cline{2-5}  
             & TrOutShaftSp               & Transmission Output Shaft Speed              & Revolutions Per Minute (rpm)            & Calculates the transmission gear ratio when in use.                                \\\cline{2-5}  \hline

 & FuelRate                   & Fuel Rate                                    & Gallons (U.S.) Per Hour (gph)             & The fuel rate is essentially how   quickly the vehicle is burning fuel.             \\\cline{2-5} 
\multirow{3}{*} {Fuel}         & InstFuelEco                & Instantaneous Fuel Economy                   & Gallons (U.S.) Per Hour (gph)             & Captures current fuel economy at current vehicle velocity.                           \\\cline{2-5}
             & InjCtlPres                 & Injector Control Pressure                    & Pound per square inch (psi)               & Can be used as a health indicator of the injectors.                                 \\\cline{2-5} \hline


\multirow{1}{*} {Brake}        & BrakeSwitch                & Brake Switch                                 & Base 10 Integer Number                  & Captures the position of break pedal.        \\\cline{2-5}     \hline     

                                    
\end{tabular}}
\\
\\
\caption{An overview of the  vehicular FWGs included in the study, along with a description of the associated sensor channels \cite{neupane2022temporal}.}

\label{table:features_description}

\vspace{-4mm}

\end{table*}

\subsection{TwinExplainer}

Understanding the ``root cause" of anomalous behavior in sensor channels is important for stakeholders in the automotive domain. It is ultimately they who make decisions such as scheduling a maintenance, 
{repairs, diagnosis, fault detection and remediation} in a timely manner that might possibly avert 
impending failure in the engine or other subsystems, and reduce the cost of maintenance. 
Therefore, the key to making a DT useful for its users is to translate information about detected/predicted 
{outcomes} into a form that can be 
easily understood by humans \cite{song2018exad}. However, this phase is typically omitted {in the body of literature}, and the majority of DT 
{systems} place a greater emphasis on prediction accuracy than on model interpretability. 

Currently, ML/DL methods are actively employed to construct 
detection {(fault, anomalies, intrusion, etc.)} systems due to their ability to attain an unprecedented detection rate. However, despite their impressive prediction accuracy, these opaque and non-transparent models cannot justify their predictions \cite{9927396}. This is due to the fact that ML/DL models have complex functions and non-linear structures, making it challenging to identify the precise data information that influences their decision-making \cite{samek2017explainable}. This lack of awareness of the inner workings of opaque AI models or the inability to retrace the outputs to the source data generates trust problems among users. \cite{marshan2021artificial}.


In this paper, we overcome precisely this issue by giving explanations for the predictions made by the deep learning-enabled DT system. Given an anomaly identified in the DT system, the mechanic or the overseer might not quickly establish the underlying sensors that are responsible for the anomaly. To overcome this problem, the \emph{TwinExplainer} system examines the feature attributions, which reflect the degree to which each feature in a model contributed to the predictions for each instance. Next, we discuss twin explainer. 

The input to this phase is the prediction made by the virtual asset or the clone of a physical system. The predictions are then provided to the explainer module, such as LIME \cite{ribeiro2016should} and SHAP\cite{lundberg2017unified}.  
The explainer module will then mimic the complex deep learning module and evaluate the feature attribution of individual predictions. The significance of local and global features is assessed. The explainer module's ability to explain a single prediction or decision is known as \textit{local feature explanation}. It is utilized to generate a unique explanation or justification for the specific decision made by the DT. On the other hand, \textit{global feature explanation} refers to the ability of the explainer module to explain the reasons behind all the possible outcomes. In this sense, such explanations will offer light on the model's decision-making process as a whole, resulting in an understanding of the attributions for varied input data and augmenting user trust in the systems.

The objective of the explainer module is to generate the system's rationale for an end user who relies on the predictions and recommendations made by DTs, or actions taken by them, in order to make a confident decision and schedule a maintenance activity, {repairs, fault diagnosis and remediation, and risk mitigation}.  An operator who monitors data-driven vehicular systems, for instance, must comprehend the system's decision-making model in order to schedule {troubleshooting, repair, and} maintenance activities effectively and in a timely manner to mitigate the risk of abrupt failure.

Next, we describe the experimental setup in Section \ref{exp} and how we use \emph{TwinExplainer} to generate explanations for predictions computed by an automotive DT in Section \ref{explanations}.

\section{Experimental Setup}\label{exp}

In this section, we describe the experimental setup that was utilized to generate the explanations. Following a description of the dataset used in the experiment, we describe our DT modeling approach.
\label{experiment}

\subsection{Vehicle Performance, Reliability, and Operations (VePRO)}\label{dataset}

In this study, we employed the {Vehicle Performance, Reliability, and Operations} 
dataset. The VePRO program is a collaboration between the United States (U.S.) Army Corps of Engineers and Mississippi State University (MSU). The detailed specifications of this dataset can be found in our previous work \cite{neupane2022temporal}. The data set is a collection of operational data of a fleet of anonymized military vehicles, and each vehicle in the dataset contains a minimum of 106 sensor channels. 

\begin{figure}[h]
    \centering
    \includegraphics[scale=.55]{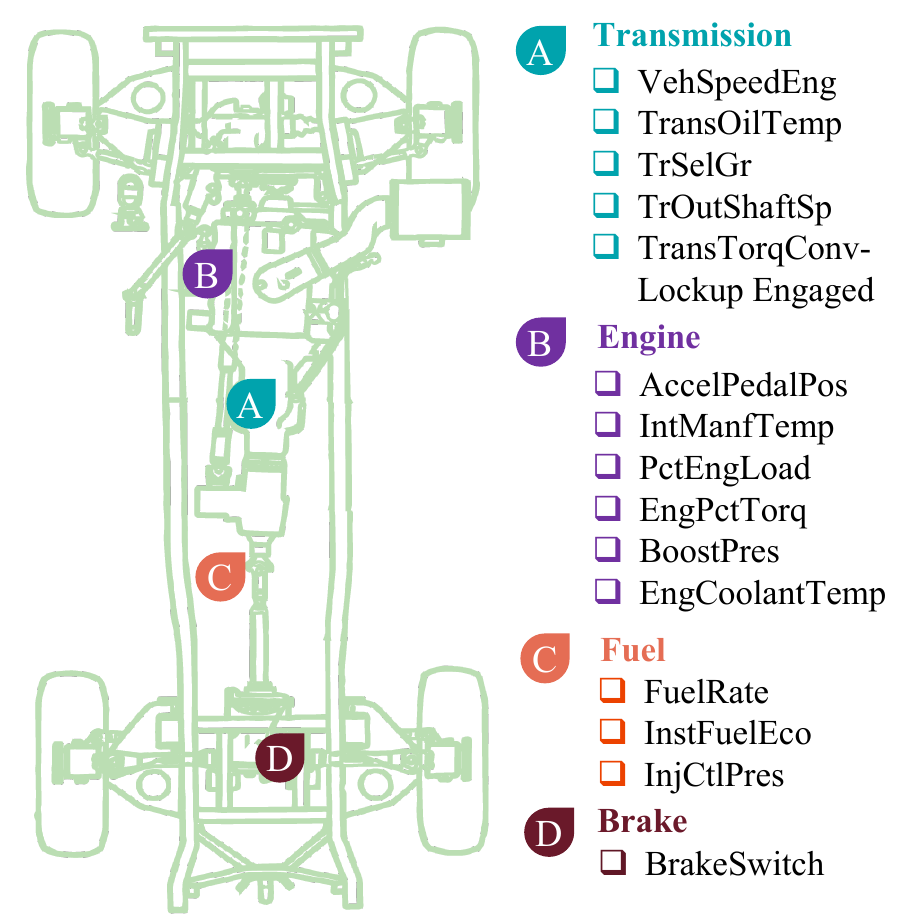}
    \caption{ A schematic representation of the engine, transmission, fuel, and braking subsystems in a vehicle, as well as their sensor channels. In total, 15 suitable sensor channels were selected based on the recommendations of mechanical engineers working on the same dataset \cite{neupane2022temporal,patterson2022white}.}
        \label{fig: subsystems}
\end{figure}

We utilize the sensor channels pertaining to the primary components of a vehicular system, such as \textit{engine, transmission, fuel}, and \emph{brakes}, to develop a DT. 
Based on the recommendation of mechanical engineers working on the same dataset, we selected 15 sensor channels relating to four subsystems (See schematic diagram Fig \ref{fig: subsystems} for related subsystems and thier channels). 
For engine subsystems, we utilize sensor channels such as \emph{engine coolant temperature, percent engine load, turbo boost pressure, accelerator pedal position, intake manifold temperature}, etc., whereas for transmission subsystems we consider \emph{vehicle speed, transmission oil temperature, transmission selected gear}, etc. For fuel subsystems, we select channels like \emph{fuel rate, injector control pressure}, etc. Likewise, for the braking subsystems, we select the brake switch sensor. 
An overview of these channels can be found in Table \ref{table:features_description}.

\begin{figure*}[h]
    \centering
    \includegraphics[scale=.60]{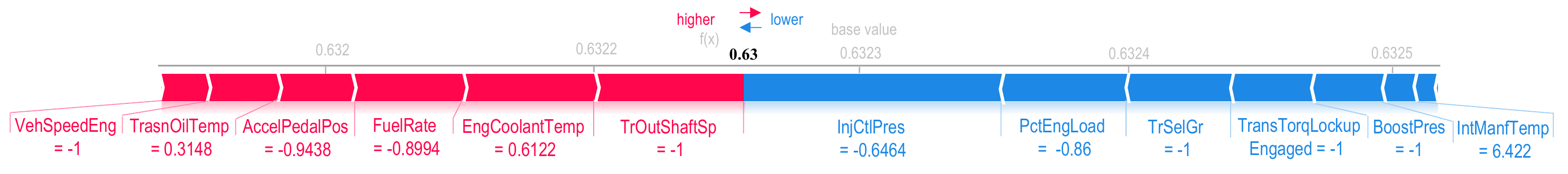}
    \caption{{A graphical depiction of a single normal prediction by DT using a force plot as a visualization tool. The red arrow in the figure represents features that positively contribute to the overall prediction, while the blue arrow represents features that negatively contribute to the prediction. Sensor channels (feature) with their respective values serve as input. Larger arrows represent SHAP-valued features with greater impact.}}
        \label{fig: force_plot_normal}
\end{figure*} 

\begin{figure*}[h]
    \centering
    \includegraphics[scale=.65]{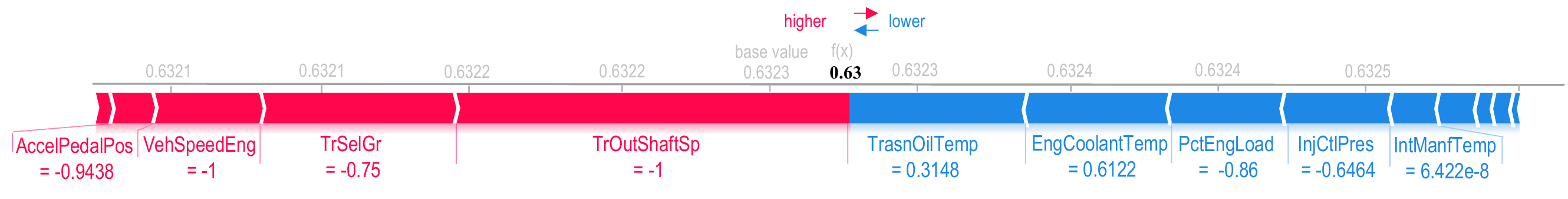}
    \caption{{Visualizing a single abnormal prediction by DT utilizing a force plot diagram. The blue arrow indicates features that are negatively contributing to the prediction, while the red arrow indicates features that are positively contributing to the prediction. The value of the feature (sensor channels) is the input as depicted in the graph. SHAP-valued features with a greater impact are represented by arrows with a larger length.}}
        \label{fig: force_plot_anomaly}
\end{figure*} 

\subsection{Modeling a Digital Twin} \label{Modeling_dt}
The existing DT model employs a two-phased approach to identify abnormalities, analogous to a one-classification task. In the first phase, predictive modeling is performed in the form of time series forecasting. The DT model ingests time series data from each sensor channel. Due to its streaming nature, the data is processed using a sliding window approach to create input sequences of contiguous observations. A Temporal Convolutional Network (TCN) is applied to predict the next observation at $t_n$ given an input sequence of observations from $t_1$ to $t_{n-1}$ (see previous work \cite{neupane2022temporal}). In the second phase, anomaly detection is performed using the TCN prediction error distribution found during training. When a query (i.e., test sequence) is made to the DT, the prediction error is computed and compared to the training prediction error distribution using the Mahalanobis distance (MD). A large MD value indicates a significant deviation from the predictive model. Using the MD, the system is able to determine if a query contains anomalous observations. 

\section{Generating Explanations Using TwinExplainer}
\label{explanations}
In this section, we describe how \emph{TwinExplainer} can be used to create explanations. {TwinExplainer} can help a user visualize both global and local explanations for predictions made by a Digital Twin (DT). 
{Bar charts, beeswarms, summary plots, dependency plots, and force plots are some of the visualization techniques employed by \emph{TwinExplainer} to illustrate the DT predictions.}
First, we'll describe global explanations, emphasizing feature attribution for a DT. This type of 
{explanation} will assist automotive stakeholders in understanding the global scale of the sensor channels and how they contribute towards generic predictions computed by the DT. Furthermore, we also formulate local explanations to assist automotive stakeholders by making it possible to comprehend the reasoning behind a specific DT prediction. {We examine two scenarios for local prediction. The first scenario entails visualizing explanations for normal predictions. In the second step of our process, we create visual representations for any abnormal or atypical predictions that the DT has produced.} 
We then make comparisons between these two scenarios and explain what features contributed most to formulating predictions for each of these scenarios. 

\subsection{Global Explanation for Digital Twin}

Global explanations strive to represent the behavior of the model as a whole, which follows the entire chain of reasoning leading to all potential outcomes. It is useful for population-level decisions. 
Therefore, global explainability is vital for knowing the overall behavior of the 
{DT} with regard to input data distributions.

\begin{figure}[h]
    
    \includegraphics[scale=.65]{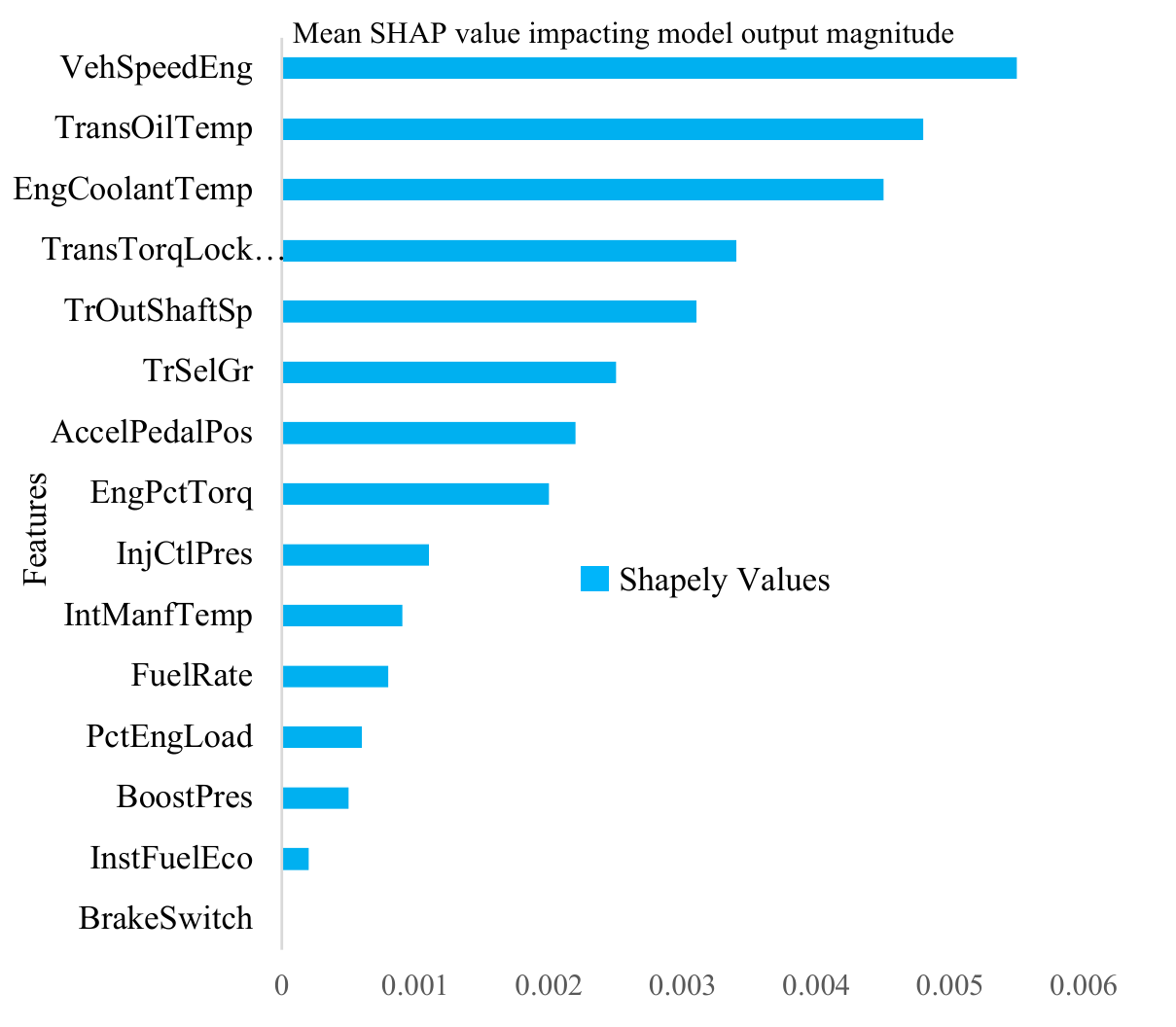}
    \caption{{The x-axis of the barchart shows the SHAP value (or contribution) for each feature of the DT, while the y-axis shows the features, ordered from most to least important.} The three most important global features are \emph{VehicleSpeedEng, TranOilTemp, and EngCoolantTemp}, with \emph{BoostPres, InstFuelEco, and BrakeSwitch} being the least important {towards prediction made by the DT.} See Table \ref{table:features_description} for channel description.}
        \label{fig: bar_plot_normal}
\end{figure}

{We utilize SHAP \cite{lundberg2017unified} plots to showcase the importance of each feature in descending order of magnitude. To demonstrate the} 
significance of global 
{features to the DT's prediction} we 
{employ two visualization tools: the bar chart} (See Fig \ref{fig: bar_plot_normal}) and the summary chart (See Fig \ref{fig: sumary_plot_normal}). In addition, we use the dependency plot (See Fig \ref{fig: dependence_plot_fuelrate}) to show interaction effects between two sensor channels.

\subsubsection{Digital Twin Feature Importance}
{To illustrate how DT arrived at its prediction, we construct the SHAP feature importance plot, which is based on the magnitude of feature attributions; thus, features with large absolute Shapley values are deemed significant.} To compute global feature importance, we average the absolute Shapley values for each feature throughout the dataset. Then, these values are used to generate a bar chart based on the contribution of each variable that are arranged in decreasing order of relevance to the prediction model. The bar chart (See Fig \ref{fig: bar_plot_normal}) depicts the relative importance of each feature, which can help automotive stakeholders gather information on the most and least influential sensor channels. In this instance, the three most influential sensor channels were \emph{VehSpeedEng, TransOilTemp}, and \emph{EngCoolantTemp}, whilst \emph{BoostPres, InstFuelEco}, and \emph{Brakeswitch} were the least influential. 

{The DT feature importance plot, while informative, does not reveal much beyond the feature's significance. The subsequent section will delve into more advanced and informative plots, such as summary plots.} 

\subsubsection{Summary Plot for Digital Twin}
\begin{figure}[h]
    
    \includegraphics[scale=.42]{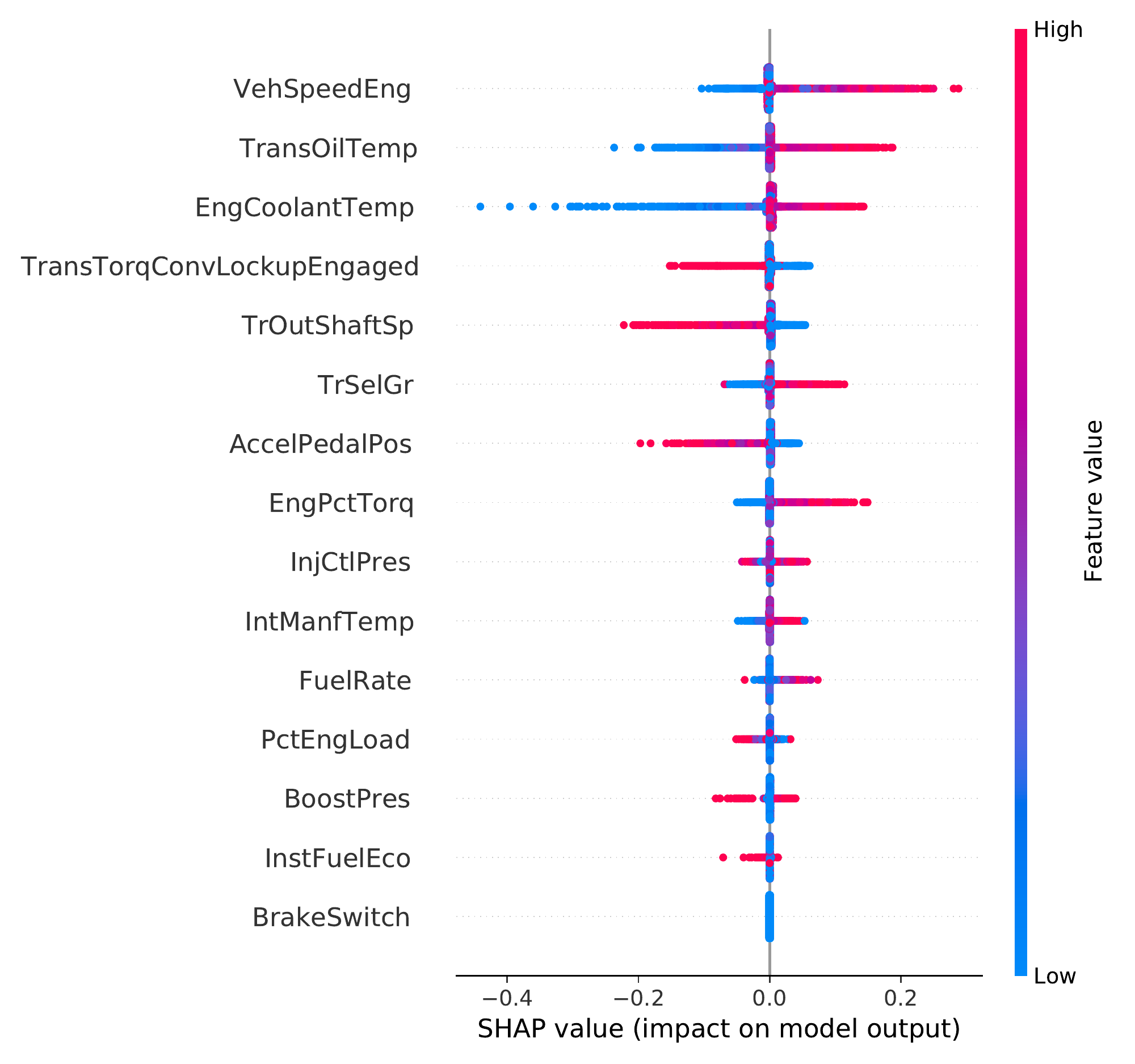}
    \caption{ {A graphical representation of a beeswarm plot that shows an information-dense summary of how the top features in a dataset influence the model's output. On each feature row, a single dot represents each instance of the given explanation. The SHAP value of that feature determines the x position of the dot, and dots pile up along each feature row to show density. The features are ordered in decreasing order based on their mean absolute SHAP values. In this case, the top three sensor channels that contribute positively towards overall prediction are} \emph{VehSpeedEng, TransOilTemp}, and \emph{EngCoolantTemp}, while the least influential sensor channels that impact prediction negatively are \emph{InstFuelEco and Brakeswitch}.}
    \label{fig: sumary_plot_normal}
\end{figure}

{The summary plot} (beeswarm chart (See Fig \ref{fig: sumary_plot_normal})) {combines feature significance and feature effects, displaying the contribution of each feature to DT prediction while accounting for all feature values, sorted by their mean absolute SHAP value, with the most significant features on top.} 
{Each point on the beeswarm graph represents a single SHAP value for a DT prediction and feature.} The horizontal axis illustrates the SHAP values of each feature utilized by the 
DT to forecast 
{predictions}. 
{The color legend on the right denotes the magnitude of features, with red representing greater feature values and blue representing lesser feature values. The distribution of red and blue points provides an indication of the directionality of features and their overall impact on DT prediction.}

Fig \ref{fig: sumary_plot_normal} represents a summary plot of our DT predictions. In this case, we can observe that most of the data points in the \emph{VehSpeedEng} sensor channel are red, which is pushing towards a positive prediction, while most points in \emph{EngCoolantTemp} are blue, which is negatively influencing the prediction. The same analysis is applicable to the remaining sensor channels. Positive and negative in this context are merely direction terms that refer to the direction in which the DT prediction is affected; they have no bearing on performance.

\begin{figure}[h]
    
    \includegraphics[scale=.45
    ]{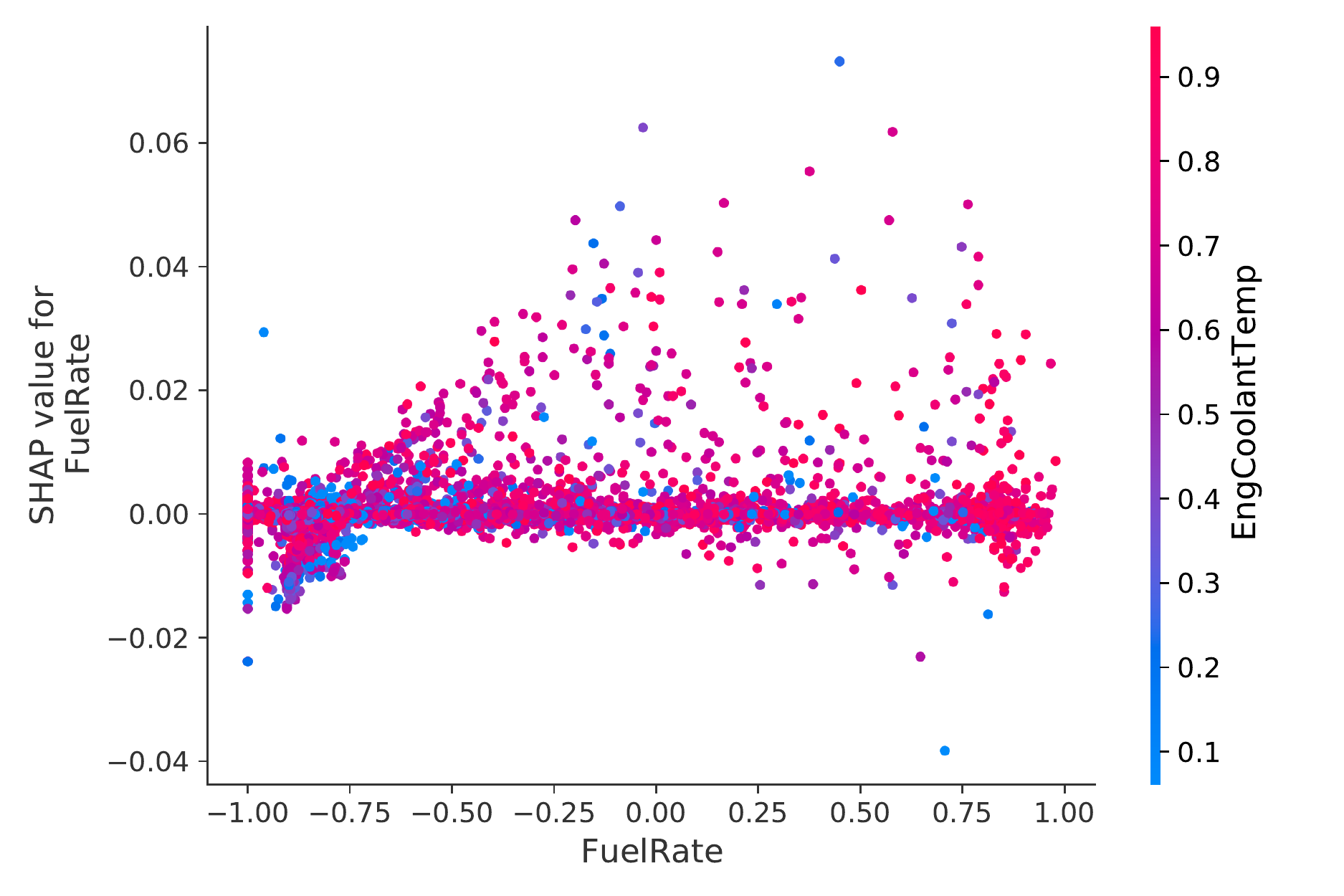}
    \caption{Partial dependence plot showcasing the sensor channel \emph{FuelRate} interacting mostly with \emph{EngCoolantTemp}. Each point on the graph represents a data instance. As \emph{FuelRate} and \emph{EngCoolantTemp} increase the overall prediction decrease. 
    }
        \label{fig: dependence_plot_fuelrate}
\end{figure} 

\subsubsection{Digital Twin Dependence Plot}
A dependence plot is a useful (See Fig \ref{fig: dependence_plot_fuelrate}) chart that shows how a single feature affects the DT predictions across the entire dataset.  Each point 
represents an individual row from the dataset. The x-axis represents the feature's value, while the y-axis represents the SHAP value for that feature. Fig \ref{fig: dependence_plot_fuelrate}, depicts a  dependence plot of sensor channels \emph{FuelRate} and \emph{EngCoolantTemp}. From the figure, it can be inferred that as the \emph{FuelRate} increases, the overall prediction score decreases. Similarly, as the \emph{EngCoolantTemp} increases, the overall prediction decreases.

\subsection{Local Explanation for Digital Twin}
The purpose of local explanation is to assist stakeholders in explaining the 
specific prediction made by the DT. This 
technique aids comprehension of individual predictions by addressing queries such as, ``How does the model’s prediction change if the input is significantly changed?" Furthermore, these explanations are based on a model that closely approximates the prediction model in the vicinity of a specific observation. Stakeholders in the automotive sector may use this strategy to learn about sensor channel characteristics and how they contribute to certain predictions. This will allow them to concentrate on channel sensors that are producing aberrant readings and offer a maintenance plan to repair problematic channels.

{Using force plots, \emph{TwinExplainer} explains the local predictions of DT for both normal and aberrant predictions. We describe the normal prediction made by the DT first, followed by the abnormal prediction. The purpose is to look into and understand what features attribution led to normal vs. abnormal predictions.}


\subsubsection{Normal Prediction}

The force plot illustrates the contribution of each feature (sensor channels) to a specific prediction computed by the DT. Fig \ref{fig: force_plot_normal}, Fig \ref{fig: force_plot_anomaly} represent force plot visualization for normal and anomalous prediction respectively. In these figures, the red arrow represents feature effects (SHAP values) that cause the prediction value to increase, whereas the blue arrow represents those that cause the prediction value to decrease. The size of each arrow represents the degree of influence of the associated feature. The average prediction of a DT throughout the training set is represented by the ``base value" (gray numbers in the top left). At the bottom of the graph, the actual feature values and it\textquotesingle s impacts are presented. 

 Fig \ref{fig: force_plot_normal} {illustrates} a force plot for a normal prediction by DT at 0.63. In this case, the input consists of all sensor channels (features), whose values are listed alongside their respective names. 
 The most important features that contributed positively to making this prediction are \emph{TrOutShaftSp, EngCoolantTemp, FuelRate, AccelPedalPos, TransOilTemp}, and \emph{VehSpeedEng} whereas, the features that contributed to negatively influence the prediction in order of their magnitude are \emph{InjCtlPres, PctEngLoad, TrSelGr, TransConvLockUpEngaged, BoostPres}, and \emph{IntManTemp}. 

\subsubsection{Abnormal Prediction} 
To explain an abnormal prediction by the DT, we employ the same 
{visualization technique}.  Fig \ref{fig: force_plot_anomaly}, is a graphical representation of a single anomalous prediction in the form of a force plot. Similar to the normal prediction, the input for this explanation consists of the same features but with different feature values, as depicted on the graph above next to their respective names. In this scenario, the contribution of characteristics differs from 
{a normal} prediction. Features such as \emph{TrOutShaftSp, TrSelGr, VehSpeedEng, AccelPedalPos} positively influenced overall prediction of 0.63 while features such as \emph{TransOilTemp, EngCoolantTemp, PctEngLoad, InjCtlPres, IntManTemp, EngPctTorq} influenced prediction negatively. The most important favorably affecting characteristic for both predictions is \emph{TrOutShaftSp} whereas the most adversely influencing feature for normal prediction is  \emph{InjCtlPres} and \emph{TransOilTemp} for abnormal prediction. 

\section{Conclusion \& Future Works}
\label{conclusion}

In light of the development of sophisticated sensing and intelligent systems, digitization will be of paramount importance in the future. Consequently, DT could be one of the most promising enabling technologies and an essential factor in determining how the cyber (virtual) and physical worlds interact with one another. In general, DT systems leverage deep learning models for analytics, insights, and business intelligence; however, the decisions of the deep learning approaches used by DT systems are not intrinsically interpretable, which creates a barrier to their implementation by raising issues of transparency in decision-making and user trust concerns. 

To resolve these issues, in this paper, we present \emph{TwinExplainer}, which explains the predictions of a DT for a vehicle. A reasonable explanation of the prediction aids automotive stakeholders in comprehending the outcomes of decisions and assisting them with downstream tasks such as predictive maintenance, fault detection and diagnosis, optimization, and repairs while augmenting their trust in AI systems. We exploited our pipeline to automate the process of generating explanations for DTs. Our pipeline has three distinct stages: starting from collecting, storing, and processing sensor channel inputs; constructing a DT utilizing a deep learning classifier that forecasts sensor channel readings; and lastly, generating explanations for the predicted outcomes. To justify prediction, \emph{TwinExplainer} employs a feature-based explainable approach. Feature-based explanations utilize features as the method of explanation to ascertain the degree to which each feature contributes to the output prediction. Two types of explanation (global and local) are provided in the form of visualization (bar charts, summary plots, dependence plots, and force plots) to illustrate the feature attribution, their influence, and their contribution to driving the model to certain decisions.

In the future, we would like to evaluate the performance of the explanations. 
In doing so, we plan to develop user-oriented evaluation metrics to investigate the impact of explanations.


\section*{Acknowledgment}
This work was supported by PATENT Lab (Predictive Analytics and TEchnology iNTegration Laboratory) at the Department of Computer Science and Engineering, Mississippi State University; a U.S. Department of Defense grant; National Science Foundation (\#1565484). The views and conclusions are those of the authors.


%

\bibliographystyle{unsrt}
\bibliography{refs}

\end{document}